\documentclass[runningheads]{llncs}
\usepackage[T1]{fontenc}
\usepackage[utf8]{inputenc}
\usepackage{floatrow}
\newfloatcommand{capbtabbox}{table}[][\FBwidth]
\usepackage[marginal]{footmisc}

\usepackage{authblk}
\usepackage{graphicx}
\usepackage{amsmath}
\usepackage{bm}
\usepackage{float}
\usepackage{makeidx}  
\usepackage{amssymb}
\usepackage{verbatim}
\usepackage{algorithm}
\usepackage{algorithmic}
\usepackage{booktabs}
\usepackage{multirow}
\usepackage{pbox} 
\usepackage{cite}
\usepackage{threeparttable}
\usepackage{caption}
\usepackage{amsfonts}
\usepackage{amssymb}
\usepackage{hyperref}
\usepackage{subcaption}
\usepackage{hyperref}
\usepackage{makecell}
\usepackage{booktabs}
\usepackage{multirow}
\usepackage{caption}
\usepackage{graphicx}

\usepackage[table]{xcolor}
\definecolor{dino}{RGB}{249,231,227}
\def\sym#1{\ifmmode^{#1}\else\(^{#1}\)\fi}

\makeatletter
\def\thanks#1{\protected@xdef\@thanks{\@thanks
        \protect\footnotetext{#1}}}
\makeatother

\begin{document}

\title{Unified Modeling Enhanced Multimodal Learning for Precision Neuro-Oncology~\thanks{Code will be available at: \href{https://github.com/HuahuiYi/MMDP}{https://github.com/HuahuiYi/MMDP}}}

\author{Huahui Yi\inst{1}$^{\dagger}$ \and Xiaofei Wang\inst{2}$^{\dagger}$ \and Kang Li\inst{1}$^{\ast}$ \and Chao Li\inst{2}$^{\ast}$ \thanks{$\ast$ Corresponding authors, $\dagger$ Equal contribution.}}
\institute{
$^1$West China Biomedical Big Data Center,West China Hospital, Sichuan University, China\\
$^2$Department of Clinical Neurosciences, University of Cambridge, UK
}
\maketitle

\begin{abstract}

Multimodal learning, integrating histology images and genomics, promises to enhance precision oncology with comprehensive views at microscopic and molecular levels. 
However, existing methods may not sufficiently model the shared or complementary information for more effective integration. 
In this study, we introduce a Unified Modeling Enhanced Multimodal Learning (UMEML) framework that employs a hierarchical attention structure to effectively leverage shared and complementary features of both modalities of histology and genomics.
Specifically, to mitigate unimodal bias from modality imbalance, we utilize a query-based cross-attention mechanism for prototype clustering in the pathology encoder. Our prototype assignment and modularity strategy are designed to align shared features and minimizes modality gaps. An additional registration mechanism with learnable tokens is introduced to enhance cross-modal feature integration and robustness in multimodal unified modeling.
Our experiments demonstrate that our method surpasses previous state-of-the-art approaches in glioma diagnosis and prognosis tasks, underscoring its superiority in precision neuro-Oncology.
\end{abstract}

\section{Introduction}\label{sec:c}

Multimodal learning~\cite{liang2022foundations} refers to the methods of integrating different types of data (e.g., images, genomics) that provides comprehensive information, promising to discover robust feature representations for disease characterization~\cite{huang2021makes}.
Integrating multimodal data is particularly relevant for characterizing cancer, a complex disease with remarkable heterogeneity. There is a pressing need to develop multimodal learning approaches for precision oncology~\cite{wei2022collaborative, wei2023multi, boehm2022harnessing}.

Histopathology is the gold standard in diagnosing cancer. The microscopic morphology observed from tissue sections provides crucial information on tumor structure and tissue compositions. Meanwhile, high-throughput sequencing has dramatically accelerated cancer discovery with in-depth genomics profiling, offering opportunities for understanding the molecular underpinnings of cancer. Integrating histopathology with genomics~\cite{wang2023multi} allows for a more holistic approach to study cancer at microscopic and molecular levels. 
Various computational approaches have been proposed to integrate genomics with histopathology features extracted from histology whole slide images (WSIs)~\cite{mobadersany2018predicting, chen2020pathomic, chen2022pan}. For instance, Mobadersany \textit{et al.}.~\cite{mobadersany2018predicting} propose to integrate features using vector concatenation, while Chen \textit{et al.}.~\cite{chen2020pathomic, chen2022pan} employ Kronecker Product to fuse features. Despite simplicity and efficiency, these methods may ignore the potential correlations and interactions between modalities. Therefore, they cannot fully leverage multimodal information to benefit from robust feature discovery.

To model cross-modal interaction, Chen \textit{et al}~\cite{chen2021multimodal} introduce Genomic-Guided Co-Attention (GCA), using genomics as guidance to identify informative WSIs instances for integration. Zhou \textit{et al.}~\cite{zhou2023cross} propose a Cross-Modal Translation and Alignment framework to transfer and integrate complementary information between modalities. Despite the success, solely relying on either shared or complementary information cannot effectively model complex biological systems for understanding cancer. Effective models remain lacking in utilizing both shared and complementary information in multimodalities. Further, due to the significant tumor heterogeneity and differences in acquisition, histopathology and genomics data present marked noise and modality gaps, challenging effective integration. 

To address these challenges, we propose a multimodal learning framework, namely Unified Modeling Enhanced Multimodal Learning (UMEML), to effectively uncover the shared and complementary features from multimodalities and integrate them for precision oncology.
Our framework is a hierarchical attention structure comprising two unimodal encoders within each modality and a unified multimodal decoder designed to decode and model the complex relations across modalities. 
WSI patches typically have much larger quantities than genomics data, causing unimodal bias~\cite{zhang2023theory}. To mitigate this challenge, we employ a query-based cross-attention mechanism in the pathology encoder, clustering patch instances into prototypes while concurrently reducing the impact of patch noise.
To effectively align shared features from multimodalities while minimizing the modality gap, we design a prototype assignment and modularity strategy.
To reduce cross-modal noise and facilitate robust multimodal modeling, a registration mechanism inspired by~\cite{darcet2023vision} is introduced, i.e., additional learnable tokens are added between multimodal prototypes, enhancing the learning process for cross-modal feature integration. 
Our contribution is fourfold: 
\begin{itemize}
    \item We propose the UMEML framework with a hierarchical attention structure. As far as we know, this is the first model to leverage the shared and complementary information between pathology and genomic modalities. 
    \item We employ a prototype query-based cross-attention mechanism in the pathology encoder to mitigate the unimodal bias and reduce patch noise. 
    \item We introduce prototype assignment and modularity strategy to effectively bridge the representation spaces and align shared features. 
    \item We introduce the registration mechanism to mitigate the noise between multimodalities, enhancing unified modeling for feature integration.   
\end{itemize}

We perform comprehensive experiments on three tasks of glioma grading, classification, and survival prediction. Results show that our model consistently outperforms other state-of-the-art methods, promising to promote precision neuro-oncology. 
\begin{figure*}[tb]
    \centering
    \includegraphics[width=0.9\linewidth]{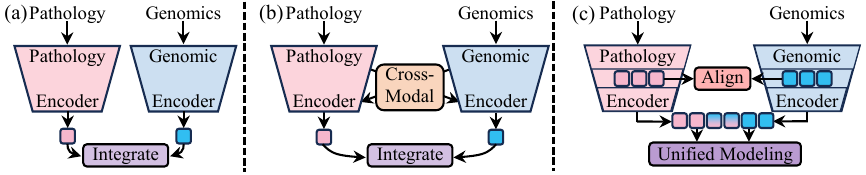}
    \caption{Illustration of multimodal fusion methods.}
    \label{fig:compare}
\end{figure*}

\section{Methodology}\label{sec:methodology}

\begin{figure*}[tb]
    \centering
    \includegraphics[width=0.85\linewidth]{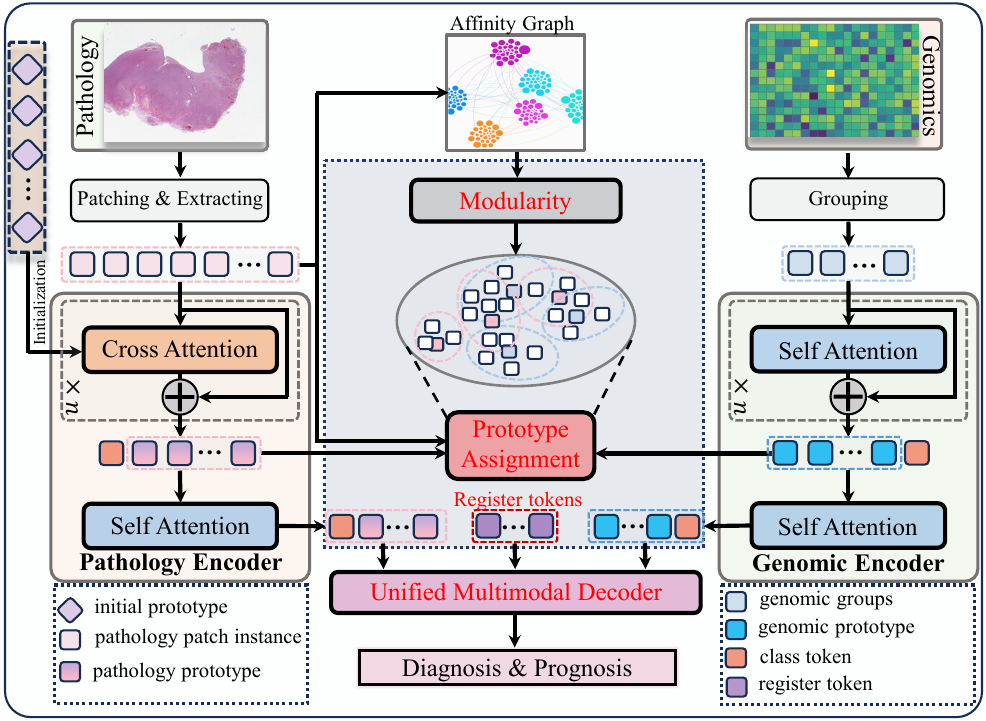}
    \caption{The proposed UMEML framework. Histopathology-genomic pairs through two unimodal encoders to derive prototypes. The Assignment and Modularity Module refines prototypes using an Affinity Graph, concatenates them with noise-mitigating registers, and inputs them into a Unified Multimodal Decoder for unified representation modeling in downstream tasks.}
    \label{fig:method}
\end{figure*}

\subsection{Problem Formulation}\label{subsec:Problem}
The primary goal of multimodal learning is to create a joint representation space that integrates information from multiple communicative modalities. 
For given pathology and genomic pair $(p, g)$ and label $y$, where $p \in \mathcal{P}$ in pathology modality, $g \in \mathcal{G}$ in genomics and $y \in \mathcal{Y}$. We denote $f_{m}$ as a deep network that maps the input in $\mathcal{M}$ modality to latent space, and features across modalities are integrated and passed to a head $h$. Training is done by minimizing the loss:
\begin{small}
\begin{align}
\mathcal{L}_{\text{multi}}=\mathcal{L}(h(f_p(p) \bigoplus f_g(g)), y) \label{formulation}
\end{align}
\end{small}
where $\bigoplus$ denotes a fusion operation. Fig~\ref{fig:compare} (a) directly adopts a straightforward fusion of pathology and genomic features (e.g., add, concatenation, Kronecker Product), ignoring the potential correlations and interactions between modalities. 
As shown in Fig~\ref{fig:compare} (b), the cross-modal relationship can be decomposed into shared and complementary information. Although studies utilized \textbf{either} information~\cite{chen2021multimodal,zhou2023cross}, we contend that, due to the complexity of cancer, it is crucial to employ \textbf{both} information simultaneously. To enhance our understanding of cancer, we propose a novel approach characterized by unified modeling, as illustrated in Fig~\ref{fig:compare} (c). Specifically, the method employs a hierarchical attention structure and an alignment module to capture shared and unique features, facilitating their unified modeling.

\subsection{Overall Structure}\label{subsec:encoder}

Our approach is a hierarchical attention structure, as shown in Fig \ref{fig:method}, beginning with two independent unimodal attention encoders, Pathology Encoder and Genomic Encoder, which model single modality, followed by a unified multimodal attention decoder that models cross-modal relationships.

\par\noindent\textbf{Pathology Encoder.}
For the given histopathology patch instance set $P = \{p_{1}, p_{2}, \cdots, p_{M}\} \in \mathbb{R}^{M \times d}$, we define a set of learnable initial prototypes that are iteratively updated through learning the affinity graph among patch instances. Additionally, given that the number of prototypes $K$ is much smaller than $M$, and approximately equal to $N$, employing cross-attention to aggregate instances into $K$ prototypes using initial prototypes as $query$ effectively mitigates unimodal bias. Specifically, we first apply cross-attention, using the initial prototypes as $query$ and the instance representations as $key$ and $value$. Let $C^l \in \mathbb{R}^{K \times d}$ denote $k$ prototypes after $l$-th update and $P \in \mathbb{R}^{M \times d}$ denote $M$ instance representations from a WSI, the cross attention can be formulated as
\begin{small}
\begin{equation}
    \bar{C^{l}} = \mathrm{Softmax}(\frac{C^{l-1} W_q (P W_k)^T}{\sqrt{d}})(X W_v), \quad C^{l} = C^{l-1} + \bar{C^{l}},
\end{equation}
\end{small}

where $ W_q, W_k, W_v\in \mathbb{R}^{d \times d} $ are learnable linear projections.
The cross-attention updates prototypes adaptively with the patch-level representations, which makes it possible to generate high-semantic centers adaptively for different WSIs.

Upon completing cross-attention processes, a learnable class token of pathology is concatenated to the front of $C^l$, forming $\hat{C^{l}}$ to be fed into a self-attention modeling the interconnection between  prototypes. $\hat{C^{l}}$ is expressed as
\begin{small}
\begin{equation}
    \hat{C^{l}} = [T_{CLS}, C^{l}].
\end{equation}
\end{small}
The self-attention updates each prototype in relation to others, thereby enhancing its awareness of other prototypes. 

\par\noindent\textbf{Genomic Encoder.}
For the given Genomics $G = \{g_{1}, g_{2}, \cdots, g_{N}\} \in \mathbb{R}^{N \times d}$, we posit that each gene group, post-grouping, serves as a natural prototype, and thus, we directly feed them into multiple self-attention modules to model the interrelations among gene groups. Similarly to  Pathology Encoder, we introduce a new learnable gene class token. This class token is concatenated with the previous output $P^l$ to form $\hat{P^{l}}$ fed into another self-attention module.

\par\noindent\textbf{Unified Multimodal Decoder.} Composed of $n$ self-attention layers. Through unified modeling, the encoder learns the shared and complementary information, thereby enhancing and augmenting the feature integration for downstream tasks.

\subsection{Prototype Assignment and Modularity}\label{subsec:assignment}

\par\noindent\textbf{Prototype Assignment.}
For given pathology prototypes $C \in \mathbb{R}^{K \times d}$, genomic prototypes $G \in \mathbb{R}^{N \times d}$, and histopathology patch instances $P \in \mathbb{R}^{M \times d}$, we calculate the respective prototype assignment matrices $S^p \in \mathbb{R}^{K \times M}$ for $C$ and $P$, $S^g \in \mathbb{R}^{N \times M}$ for $G$ and $P$. Taking $S^p$ as an example, the equation is as follows:
\begin{small}
\begin{equation}
    S^p = \max(0, \cos<C, P>),
\end{equation}
\end{small}
Thereby, we allocate a certain number of patch instances to each prototype and segment a WSI into $K$ and $N$ regions based on pathology and genomics prototypes. Each region, defined by a unique concept, contains associated instances.

\par\noindent\textbf{Modularity.}
We employ modularity to optimize the assignment between prototypes and patches, a method commonly used in community detection ~\cite{newman2004finding, newman2006modularity, li2023acseg}. Specifically, we first construct a fully connected, undirected affinity graph $A \in \mathbb{R}^{M \times M}$ for instances from a WSI, treated as vertices. The expression for $A$ is:
\begin{small}
\begin{equation}
    A = \max(0, \cos<P, P>),
\end{equation}
\end{small}
We then compute a weight matrix $W$ to estimate the intensity of assigning them to the same concept by
\begin{small}
\begin{equation}
    W = A-\frac{dd^{\text{T}}}{2e},
\end{equation}
\end{small}
where degree vector $d \in \mathbb{R}^{M}$ denotes the number of connected edges in its affinity $A$, and $e \in \mathbb{R}$ signifies the total edge number. By considering both prototype assignment $S$ and affinity matrix $A$, the modularity loss is formulated as:
\begin{small}
\begin{equation}
   \mathcal{L}_{\rm modularity} = -\frac{1}{2e}(\alpha \text{Tr}(W({S^p}^TS^p))+ \beta\text{Tr}W({S^g}^TS^g))),
\label{eqn:modularity}
\end{equation}
\end{small}
Here, $\alpha$  and $\beta$ are hyperparameters that modulate the impacts of various components within the modularity loss function, respectively. This loss refines prototype similarity for various instance pairs by assessing their likelihood of sharing a concept, thereby updating the prototype locations in the representation space while considering instance-specific locality and connectivity. Then the total loss is calculated as 
\begin{small}
\begin{equation}
      \mathcal{L}_{\text{total}} = \mathcal{L}_{\rm objective} + \gamma\mathcal{L}_{\rm modularity},
\end{equation}
\end{small}
Here, $\gamma $ represents a positive hyperparameter that balances the impact of the modularity loss function on the overall model performance. For objective part, We use the cross-entropy loss for grading and classification and negative log-likelihood survival loss~\cite{chen2021multimodal} for survival prediction.

\subsection{Registration Mechanism}\label{subsec:decoder}
The registration mechanism~\cite{darcet2023vision} adds a set of additional learnable tokens, called register tokens, between pathology and genomic prototypes. By learning additional relationships during the attention process, this mechanism helps reduce the disturbances caused by outliers and mismatches. Concretely, for the given pathology prototypes $\hat{C}^{l+1}\in \mathbb{R}^{(K+1) \times d}$, genomic prototypes $\hat{G}^{l+1}\in \mathbb{R}^{(N+1) \times d}$, we introduce additional learnable register tokens $R \in \mathbb{R}^{I \times d}$. These elements are then concatenated,
\begin{small}
\begin{equation}
     \hat{U} = [\hat{C}^{l+1}, R, \hat{G}^{l+1}].
\end{equation}
\end{small}
The concatenated output $\hat{U}$ is fed into the unified multimodal decoder.

\section{Experiments}\label{sec:experiments}

\subsection{Datasets and Experiments Setting}

\par\noindent\textbf{Dataset.}
We evaluate our model on diagnosis (grading, classification) and prognosis (survival prediction) using the TCGA GBM-LGG dataset~\cite{tcga}, comprising 939 patient samples and 1831 WSIs after excluding low-quality WSIs or those missing labels following prior work ~\cite{lu2021data, wang2023multi}. We crop each WSI into non-overlapping $224\text{px} \times 224\text{px}$ patches at $0.5 \mu \text{m px}^{-1}$. 
CLIP ViT-B/16~\cite{clip} is used to extract 512-dimensional features from histopathology patches, and the top-K genes with the highest expression variance are selected and categorized into $N$ uniform groups for genomic profile input.

\par\noindent\textbf{Experiments Setting.}
All experiments were conducted on a single NVIDIA RTX 3090 GPU using the PyTorch library~\cite{paszke2019pytorch} within the Python environment, with a batch size 1. To ensure representative experimental results, we employ a 5-fold cross-validation approach, using the corresponding seed for each fold during training  (e.g., $\text{fold}=1$, $\text{seed=1}$). Experiments used SGD with a 1e-5 weight decay, training the model for ten epochs with a 1e-3 learning rate for diagnosis and five epochs with a 2e-4 learning rate for prognosis. Results were averaged over five folds, considering each fold's final epoch.  

\begin{table*}[t]
 \scriptsize
   \begin{center}
      \begin{tabular}{lccccccc}
         \toprule
         
         \multirow{2}{*}{Methods} 
         &\multicolumn{2}{c}{Modality} 
         &\multicolumn{2}{c}{Grading}   
         &\multicolumn{2}{c}{Classification} 
         &\multicolumn{1}{c}{Survival}  \\
         
         \cmidrule(lr){2-3} 
         \cmidrule(lr){4-5}  
         \cmidrule(lr){6-7} 
         \cmidrule(lr){8-8}
         
          & P.    & G.  
          & Acc   & AUC    
          & Acc   & AUC    
          & C-Index \\
          
         \midrule

         SNN~\cite{klambauer2017self}      
         &     & $\checkmark$  
         &0.7054\textsuperscript{$\pm$ 0.0187}     
         &0.8438\textsuperscript{$\pm$ 0.0205}     
         &0.5956\textsuperscript{$\pm$ 0.0345}     
         &0.8476\textsuperscript{$\pm$ 0.0246}
         &0.7746\textsuperscript{$\pm$ 0.0413}\\
         
         AttMIL~\cite{ilse2018attention}   
         &$\checkmark$     &   
         &0.6392\textsuperscript{$\pm$ 0.0229}     
         &0.8118\textsuperscript{$\pm$ 0.0167}     
         &0.5482\textsuperscript{$\pm$ 0.0315}     
         &0.7998\textsuperscript{$\pm$ 0.0274}
         &0.7560\textsuperscript{$\pm$ 0.0434}\\
         
         TransMIL~\cite{shao2021transmil}   
         &$\checkmark$     &   
         &0.5852\textsuperscript{$\pm$ 0.0206}     
         &0.7420\textsuperscript{$\pm$ 0.0126}     
         &0.4584\textsuperscript{$\pm$ 0.0233}     
         &0.7068\textsuperscript{$\pm$ 0.0210}
         &0.7110\textsuperscript{$\pm$ 0.0361}\\

         Add   
         &$\checkmark$     & $\checkmark$   
         &0.6924\textsuperscript{$\pm$ 0.0415}     
         &0.8598\textsuperscript{$\pm$ 0.0161}     
         &0.6072\textsuperscript{$\pm$ 0.0191}     
         &0.8710\textsuperscript{$\pm$ 0.0275}
         &0.7960\textsuperscript{$\pm$ 0.0514}\\
         
         Concat       
         &$\checkmark$     & $\checkmark$  
         &0.7264\textsuperscript{$\pm$ 0.0406}     
         &0.8864\textsuperscript{$\pm$ 0.0164}     
         &0.6740\textsuperscript{$\pm$ 0.0160}     
         &0.9218\textsuperscript{$\pm$ 0.0100}
         &0.8300\textsuperscript{$\pm$ 0.0324}\\

         Kronecker       
         &$\checkmark$     & $\checkmark$  
         &0.7312\textsuperscript{$\pm$ 0.0401}     
         &0.8830\textsuperscript{$\pm$ 0.0175}     
         &0.6596\textsuperscript{$\pm$ 0.0353}     
         &0.9096\textsuperscript{$\pm$ 0.0181}
         &0.8224\textsuperscript{$\pm$ 0.0291}\\
         
         MCAT~\cite{chen2021multimodal}                 
         &$\checkmark$     & $\checkmark$  
         &\color{blue}0.7376\textsuperscript{$\pm$ 0.0307}     
         &\color{blue}0.9092\textsuperscript{$\pm$ 0.0164}     
         &0.6932\textsuperscript{$\pm$ 0.0480}     
         &0.9470\textsuperscript{$\pm$ 0.0113}
         &\color{blue}0.8352\textsuperscript{$\pm$ 0.0227}\\
         
         HFBSurv~\cite{li2022hfbsurv}
         &$\checkmark$     & $\checkmark$   
         &0.6906\textsuperscript{$\pm$ 0.0193}     
         &0.8506\textsuperscript{$\pm$ 0.0142}     
         &0.6600\textsuperscript{$\pm$ 0.0239}     
         &0.8506\textsuperscript{$\pm$ 0.0244}
         &0.8202\textsuperscript{$\pm$ 0.0307}\\
         
         CMAT~\cite{zhou2023cross}
         &$\checkmark$     & $\checkmark$  
         &0.7228\textsuperscript{$\pm$ 0.0226}     
         &0.9020\textsuperscript{$\pm$ 0.0175}     
         &\color{blue}0.7186\textsuperscript{$\pm$ 0.0289}     
         &\color{blue}0.9538\textsuperscript{$\pm$ 0.0070}
         &0.8286\textsuperscript{$\pm$ 0.0383}\\
         
         \rowcolor{dino}  (Ours)      
         &$\checkmark$     & $\checkmark$  
         &\color{red}0.7756\textsuperscript{$\pm$ 0.0178}     
         &\color{red}0.9212\textsuperscript{$\pm$ 0.0147}     
         &\color{red}0.7514\textsuperscript{$\pm$ 0.0380}     
         &\color{red}0.9594\textsuperscript{$\pm$ 0.0069}
         &\color{red}0.8396\textsuperscript{$\pm$ 0.0292}\\ 
         
         \bottomrule
      \end{tabular}
   \end{center}
   \caption{The performance of different approaches on three common medical tasks on TCGA GBM-LGG dataset. ``P.'' indicates whether pathological images are used and ``G.'' indicates whether genomic profiles are used. The best and second best results are highlighted in {\color{red} red} and {\color{blue} blue}, respectively.}
   \label{tab:main}
\end{table*}
\subsection{Performance Evaluation}
We implemented and compared state-of-the-art methods and baselines, covering single-modal  (SNN~\cite{klambauer2017self}, AttMIL~\cite{ilse2018attention}, TransMIL~\cite{shao2021transmil}) and multimodal learning  (Add, Concat, Kronecker Product, MCAT~\cite{chen2021multimodal}, HFBSurv~\cite{li2022hfbsurv}, CMAT~\cite{zhou2023cross}) learning approaches. Table~\ref{tab:main} shows the results of all methods. 
\par\noindent\textbf{Compared with Single-modal Models.}
As shown in Table~\ref{tab:main}, our proposed method consistently outperforms other models in all tasks. For grading, our model achieves an accuracy of 77.56\% and an AUC of 92.12\%, surpassing the best single-modal method by 7.02\% and an AUC of 7.74\%, respectively. For classification, it attains an accuracy of 75.14\% and an AUC of 95.94\%, exceeding the best single-modal method by 15.58\% and 11.18\%, respectively. For survival prediction, our model achieves a c-index of 83.96\%, an improvement of 6.50\% over the top single-modal method. These results underscore the benefit of multimodal learning, where integrating multimodalities may enhance model performance.

\par\noindent\textbf{Compared with Multimodal Models.}
For grading, Table~\ref{tab:main} (columns 4 and 5) demonstrates that our method outperforms the state-of-the-art multimodal method, MCAT, surpassing the accuracy by 3.80\% and AUC by 1.20\%. For classification, our approach surpasses the leading method, CMAT (columns 6 and 7), with an increase of 3.28\% in accuracy and 0.56\% in AUC. Furthermore, for survival prediction, a 0.44\% improvement in the c-index over MCAT is shown in column 8. These results suggest our model's advantages of leveraging shared and complementary information. Our method also significantly outperforms other comparison multimodal methods, including Add, Concat, Kronecker Product, and HFBSurv~\cite{li2022hfbsurv}, demonstrating consistently superior performance.

\subsection{Ablation Studies}
We remove or replace the key components to investigate their impact on model performance in (Table~\ref{tab:ablation}). 
After removing the modularity loss, the accuracy for grading decreases by 1.00\% and the AUC by -0.04\%. For classification, the accuracy drops by 1.40\% and the AUC by 0.16\%. For survival prediction, the c-index reduces by 0.28\%. These results imply that minimizing the modality gap benefits unified multimodal modeling. 
When the Unified Multimodal Decoder (UMD) is replaced by Bi-fusion, a widely-used cross-attention-based multimodal integration module, the accuracy for grading decreases by 1.44\% and the AUC by 0.20\%. For classification, the accuracy drops by 2.48\% and the AUC by 0.56\%. For survival prediction, the c-index reduces by 0.20\%. These findings imply that compared to Bi-fusion, UMD can model more complex relationships and suit multimodal cancer modeling.
After removing the register, we observe a decrease in accuracy by 1.36\% and AUC by 0.44\% for grading, a decrease in accuracy by 2.48\% and AUC by 0.48\% for classification, and a decrease in the c-index by 0.28\% for survival prediction. These results imply that the register contributes to the UMD's ability to model more robust multimodal relationships.

\begin{figure*}[tb]
    \centering
    \includegraphics[width=0.9\linewidth]{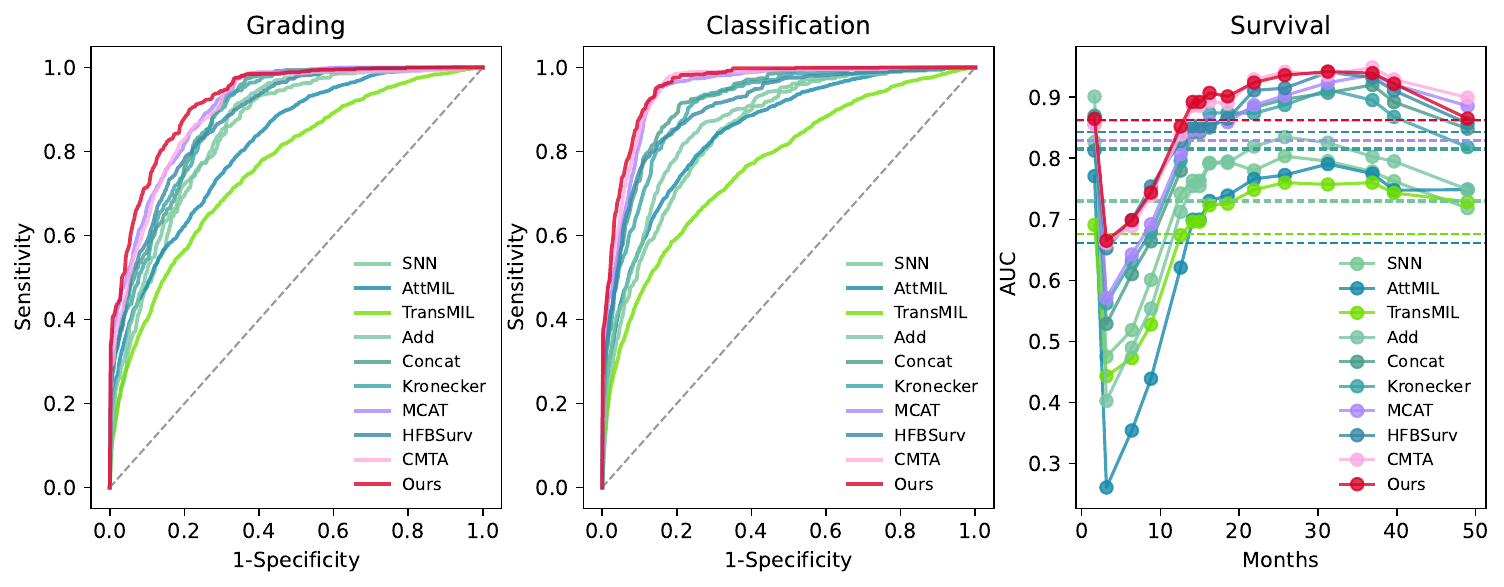}
    \caption{Compare our method's ROC curves for glioma grading and classification and time-dependent AUC curves for survival prediction against other methods.}
    \label{fig:roc}
\end{figure*}

Additionally, we have illustrated both ROC curves (glioma grading and classification) and time-dependent AUC curves (survival prediction) in Fig~\ref{fig:roc} further to demonstrate the superiority and robustness of our method.
\begin{table*}[t]
 \scriptsize
   \begin{center}
      \begin{tabular}{lccccc}
         \toprule
         {\multirow{2}{*}{Components}} 
         &\multicolumn{2}{c}{Grading}   
         &\multicolumn{2}{c}{Classification} 
         &\multicolumn{1}{c}{Survival}  \\
         
         \cmidrule(lr){2-3} 
         \cmidrule(lr){4-5}  
         \cmidrule(lr){6-6}

          & Acc   & AUC    
          & Acc   & AUC    
          & C-Index \\
          
         \midrule
         
         w/o Modularity       
         &0.7656\textsuperscript{$\pm$ 0.0175}     
         &\color{red}0.9216\textsuperscript{$\pm$ 0.0142}     
         &0.7374\textsuperscript{$\pm$ 0.0305}     
         &0.9578\textsuperscript{$\pm$ 0.0072}
         &0.8368\textsuperscript{$\pm$ 0.0252}\\
         
         w/o UMD  
         &0.7612\textsuperscript{$\pm$ 0.0216}     
         &0.9192\textsuperscript{$\pm$ 0.0187}     
         &0.7266\textsuperscript{$\pm$ 0.0315}     
         &0.9538\textsuperscript{$\pm$ 0.0074}
         &0.8376\textsuperscript{$\pm$ 0.0226}\\
         
         w/o Registers           
         &0.7620\textsuperscript{$\pm$ 0.0195}     
         &0.9168\textsuperscript{$\pm$ 0.0212}     
         &0.7266\textsuperscript{$\pm$ 0.0216}     
         &0.9546\textsuperscript{$\pm$ 0.0061}
         &0.8368\textsuperscript{$\pm$ 0.0288}\\
  
         \rowcolor{dino}  Ours (All)      
         &\color{red}0.7756\textsuperscript{$\pm$ 0.0178}     
         &0.9212\textsuperscript{$\pm$ 0.0147}     
         &\color{red}0.7514\textsuperscript{$\pm$ 0.0380}     
         &\color{red}0.9594\textsuperscript{$\pm$ 0.0069}
         &\color{red}0.8396\textsuperscript{$\pm$ 0.0292}\\ 
         
         \bottomrule
      \end{tabular}
   \end{center}
   \caption{The ablation study of three key components: 1) Omitting the Modularity Loss; 2) Replacing the Unified Multimodal Decoder (UMD) with a Bi-fusion module; 3) Excluding the using of registers in the UMD.}
   \label{tab:ablation}
\end{table*}

\section{Conclusion}\label{sec:conclusion}

This paper proposes a unified modeling approach with a hierarchical attention structure that effectively leverages both shared and complementary information from multimodal learning. To avoid unimodal bias, we address the imbalance between pathology patches and genomics groups through prototype-based cross-attention. We align these modalities using prototype assignment and modularity to reduce modeling errors arising from the modality gap. This Unified Multimodal Decoder is supplemented by the registration mechanism to reduce noises between modalities. The results show that our model achieves state-of-the-art performance. Given the cost of acquiring multimodal data, future research will address missing modalities in multimodal learning.

\newpage

\bibliographystyle{splncs}
\bibliography{arxiv.bib}

\end{document}